\documentclass[sigconf,manuscript]{acmart}
\usepackage{multirow}

\newcommand{\mE}{\mathbb{E}}

\newcommand{\oneY}{Y(1)}
\newcommand{\zeroY}{Y(0)}

\usepackage{algorithm}
\usepackage{algpseudocode}

\AtBeginDocument{%
  \providecommand\BibTeX{{%
    \normalfont B\kern-0.5em{\scshape i\kern-0.25em b}\kern-0.8em\TeX}}}

\setcopyright{acmcopyright}
\copyrightyear{2018}
\acmYear{2018}
\acmDOI{10.1145/1122445.1122456}

\acmConference[Woodstock '18]{Woodstock '18: ACM Symposium on Neural
  Gaze Detection}{June 03--05, 2018}{Woodstock, NY}
\acmBooktitle{Woodstock '18: ACM Symposium on Neural Gaze Detection,
  June 03--05, 2018, Woodstock, NY}
\acmPrice{15.00}
\acmISBN{978-1-4503-XXXX-X/18/06}

\begin{document}
\fancyhead{}
\title{
Data-Driven Off-Policy Estimator Selection:\\
An Application in User Marketing on An Online Content Delivery Service
}

\author{Yuta Saito}
\authornote{Both authors contributed equally to the paper.}
\affiliation{CFMLab.}
\email{usaito98@gmail.com}

\author{Takuma Udagawa}
\authornotemark[1]
\affiliation{Sony Corporation}
\email{Takuma.Udagawa@sony.com}

\author{Kei Tateno}
\affiliation{Sony Corporation}
\email{Kei.Tateno@sony.com}

\renewcommand{\shortauthors}{Y. Saito.}




\maketitle

\section{Introduction}
\textit{Off-policy evaluation} (OPE) is the method that attempts to estimate the performance of decision making policies using historical data generated by different policies without conducting costly online A/B tests~\cite{dudik2011doubly,gilotte2018offline}.
Accurate OPE is essential in domains such as healthcare, marketing or recommender systems to avoid deploying poor performing policies, as such policies may hart human lives or destroy the user experience.
Thus, many OPE methods with theoretical backgrounds have been proposed, including \textit{Direct Method} (DM), \textit{Inverse Probability Weighting} (IPW), and \textit{Doubly Robust} (DR).
One emerging challenge with this trend is that a suitable estimator can be different for each application setting.
For example, DM has low variance but has a large bias, and thus, performs better in small sample settings. 
On the other hand, IPW has a low bias but has a large variance, and thus reveals better performance in large sample settings.
It is often unknown for practitioners which estimator to use for their specific applications and purposes.
To find out a suitable estimator among many candidates, we use a data-driven estimator selection procedure for off-policy policy performance estimators as a practical solution.
As proof of concept, we use our procedure to select the best estimator to evaluate coupon treatment policies on a real-world online content delivery service.
In the experiment, we first observe that a suitable estimator might change with different definitions of the outcome variable, and thus the accurate estimator selection is critical in real-world applications of OPE.
Then, we demonstrate that, by utilizing our estimator selection procedure, we can easily find out suitable estimators for each purpose.
We believe that our estimator selection procedure and case study help practitioners identify the best OPE method for their environments.

\section{Setup and Method}
We denote $X \in \mathcal{X}$ as a context vector and $T \in \mathcal{T} = \{0, 1\}$ as a binary treatment assignment indicator\footnote{Note that our OPE procedure can easily be extended to multiple treatment cases.}. 
When an individual user $i$ receives the treatment, $T_i = 1$, otherwise, $T_i = 0$. 
We assume that there exist two potential outcomes denoted as $(\zeroY, \oneY)$ for each individual. 
$\zeroY$ is a potential outcome associated with $T=0$, and $\oneY$ is associated with $T=1$. 
Note that each individual receives only one treatment, and only a potential outcome for the received treatment is observed.
We can represent the observed outcome as: $Y = T \oneY + (1-T) \zeroY $. 

A \textit{policy} automatically assigns treatments to users aiming to maximize the outcome.
We denote a policy as a function that maps a context vector to one of the possible treatments, i.e., $\pi: \mathcal{X} \rightarrow \mathcal{T}$.
Then, the performance of a policy is defined as $ V (\pi) = \mE_{(X, Y)} [ Y (\pi (X)) ] $.
The goal of OPE is to estimate $ V (\pi) $ for a given \textbf{new} policy $\pi$ using log data $\mathcal{D} = \{ (X_i, T_i, Y_i) \}_{i=1}^{n}$ collected by an \textbf{old} policy different from $\pi$.

Our strategy to select the suitable estimator is to use two sources of logged bandit feedback collected by two different behavior policies.
We denote log data generated by $\pi_A$ and $\pi_B$ as $\mathcal{D}_A = \{ (X^A_i, T^A_i, Y^A_i) \}_{i=1}^{n_A}$ and $\mathcal{D}_B = \{ (X^B_i, T^B_i, Y^B_i) \}_{i=1}^{n_B}$, respectively.
To evaluate the performance of an estimator $\hat{V}$, we first estimate policy performances of $\pi_A$ and $\pi_B$ by $\hat{V}(\pi_A; \mathcal{D}_B)$ and $\hat{V}(\pi_B; \mathcal{D}_A)$.
Then, we use \textit{on-policy} estimates as the ground-truth policy performances, i.e., $ V(\pi_A) \approx \hat{V} (\pi_A; \mathcal{D}_A) = n_A^{-1} \sum_{i=1}^{n_A} Y^A_i$ and $ V(\pi_B) \approx \hat{V} (\pi_B; \mathcal{D}_B) = n_B^{-1} \sum_{i=1}^{n_B} Y^B_i$.
Finally, we compare the off-policy estimates $\hat{V}(\pi_A; \mathcal{D}_B)$ and $\hat{V}(\pi_B; \mathcal{D}_A)$ with their ground-truths (on-policy estimates) $V(\pi_A)$ and $V(\pi_B)$ to evaluate the estimation accuracy of an estimator $\hat{V}$.
We evaluate the estimation accuracy of an estimator $\hat{V}$ by the \textit{relative root mean-squared-error} defined as $\sqrt{ K^{-1} \sum_{k=1}^K (\frac{ \hat{V} (\pi_A; \mathcal{D}_B^{(k)}) - V(\pi_A) }{V(\pi_A)})^{2} }$ (same for $\pi_B$) where $k$ denotes a different subsample of logged bandit feedback made by the sample splitting or bootstrap sampling.
By applying the above procedure to several candidate estimators, we can select the estimator having the best estimation accuracy among candidates in a data-driven manner.

\begin{table*}[t]
   \begin{minipage}[t]{.48\textwidth}
    \centering
    \caption{Relative RMSE of off-policy estimators when $Y$ is \textbf{content consumption indicator}}
    \scalebox{.9}{
    \begin{tabular}{ccccc} 
    \toprule
     && \multicolumn{3}{c}{OPE Estimators} \\\cmidrule{3-5}
    OPE Situation && DM & IPW & DR \\ \midrule \midrule
    $ \mathcal{D}_A \rightarrow \pi_B $ && \textbf{0.0897} & 0.1958 & 0.0955  \\
    $ \mathcal{D}_B \rightarrow \pi_A $ && \textbf{0.0653} & 0.1589 & 0.0798 \\
    \bottomrule
    \end{tabular}}
  \end{minipage}
  \hfill
  \begin{minipage}[t]{.48\textwidth}
    \centering
    \caption{Relative RMSE of off-policy estimators when $Y$ is \textbf{revenue from users}}
    \scalebox{.9}{
    \begin{tabular}{ccccc} 
    \toprule
     && \multicolumn{3}{c}{OPE Estimators} \\\cmidrule{3-5}
    OPE Situation && DM & IPW & DR \\ \midrule \midrule
    $ \mathcal{D}_A \rightarrow \pi_B $ && 0.2231 & 0.1997 & \textbf{0.1118}  \\
    $ \mathcal{D}_B \rightarrow \pi_A $ && 0.4382 & \textbf{0.0981} & 0.2936 \\
    \bottomrule
    \end{tabular}}
  \end{minipage}
\vskip 0.05in
\raggedright
\fontsize{8.5pt}{9.5pt}\selectfont \textit{Note}: 
$ \mathcal{D}_A \rightarrow \pi_B $ is a case where we attempt to estimate the performance of $\pi_B$ using log data generated by $\pi_A$.
In contrast, $ \mathcal{D}_B \rightarrow \pi_A $ is a case where we attempt to estimate the performance of $\pi_A$ using log data generated by $\pi_B$.
The \textbf{bold fonts} represent the best off-policy estimator among DM, IPW, and DR for each setting (lower value is better).
\end{table*}

\section{A Case Study}
To show the usefulness of our procedure, we constructed $\mathcal{D}_A$ and $\mathcal{D}_B$ by randomly assigning two different policies ($\pi_A$ and $\pi_B$) to users on our content delivery platform.
Here, $X$ is a user's context vector, $T$ is a coupon assignment indicator, and $Y$ is either a user's content consumption indicator (binary) or the revenue from each user (continuous). 

We report the estimator selection results for each definition of $Y$ in Table 1 and 2.
We used DM, IPW, and DR as candidate off-policy estimators.
The tables show that different estimators should be used for each setting and purpose.
This is because the prediction accuracy of the outcome regressor used in DM and DR can be different for each definition of $Y$.
We conclude from the results that we should use DM when we want to maximize the users' content consumption probability. 
In contrast, we use IPW or DR when we consider the revenue from users as the outcome.
After the successful empirical verification, our data-driven estimator selection method has been used to decide which estimators to use to create coupon allocation policies on our platform.

\bibliographystyle{ACM-Reference-Format}
\bibliography{ref}

\end{document}